\documentclass{IEEEtran}

\pdfoutput=1

\usepackage{graphicx,amsmath,array,booktabs,multicol,calc}
\usepackage[autolinebreaks,useliterate]{mcode}
\usepackage{pst-pdf}

\renewcommand*\descriptionlabel[1]{\hspace\labelsep                                 \normalfont\itshape #1}

\newcommand{\Bernoulli}{\mathrm{Bernoulli}}
\newcommand{\BetaInc}{\mathrm{BetaInc}}
\newcommand{\Poisson}{\mathrm{Poisson}}
\newcommand{\Normal}{\mathrm{Normal}}
\newcommand{\Dirichlet}{\mathrm{Dirichlet}}

\newcommand{\CRP}{\mathrm{CRP}}
\newcommand{\IBP}{\mathrm{IBP}}
\newcommand{\GFT}{\mathrm{GFT}}
\newcommand{\UT}{\mathrm{UT}}
\newcommand{\UBT}{\mathrm{UBT}}
\newcommand{\Beta}{\mathrm{Beta}}
\newcommand{\B}{\mathrm{B}}
\newcommand{\dd}{\mathrm{d}}

\newcommand{\matlab}{MATLAB}

\begin{document}
\title{Non-parametric Bayesian modeling of complex networks}
\author{Mikkel N. Schmidt and Morten M{\o}rup}
\date{Section for Cognitive Systems, DTU Informatics, Technical University of Denmark}
\maketitle
\begin{abstract}
  Modeling structure in complex networks using Bayesian non-parametrics makes it possible to specify flexible model structures and infer the adequate model complexity from the observed data.  This paper provides a gentle introduction to non-parametric Bayesian modeling of complex networks: Using an infinite mixture model as running example we go through the steps of deriving the model as an infinite limit of a finite parametric model, inferring the model parameters by Markov chain Monte Carlo, and checking the model's fit and predictive performance. We explain how advanced non-parametric models for complex networks can be derived and point out relevant literature.
\end{abstract}

\section{Introduction}
We are surrounded by complex networks. From the networks of cell interaction in our immune system to the complex network of neurons communicating in our brain, our cells signal to each other to coordinate the functions of our body. We live in cities with complex power and water systems and these cities are linked by advanced transportation systems. We interact within social circles and our computers are connected through the Internet forming the World Wide Web. To understand the structure of these large systems of biological, physical, social, and virtual networks, there is a great need to be able to model them mathematically~\cite{borner2007network}.

Complex networks are studied in several different fields from computer science and engineering to physics, biology, sociology, and psychology. ``Network science is an emerging, highly interdisciplinary research area that aims to develop theoretical and practical approaches and techniques to increase our understanding of natural and manmade networks''~\cite{borner2007network}. Network science can be considered ``the study of network representations of physical, biological, and social phenomena leading to predictive models of these phenomena''~\cite{national2005Network}.

To understand the many large-scale complex networks we sample and store today, there is a growing demand for advanced mathematical and statistical models that can account for the structure in these systems. The modeling aims are twofold; to provide a comprehensible description (i.e., descriptive modeling) and to infer unobserved properties (i.e., predictive modeling). In particular, a statistical analysis is useful when the focus lies beyond single node properties and local interactions but on the characteristics and behaviors of the entire system~\cite{borner2007network,leskovec2008dynamics,sporns2010networks}.

\begin{figure*} 
  \centering
  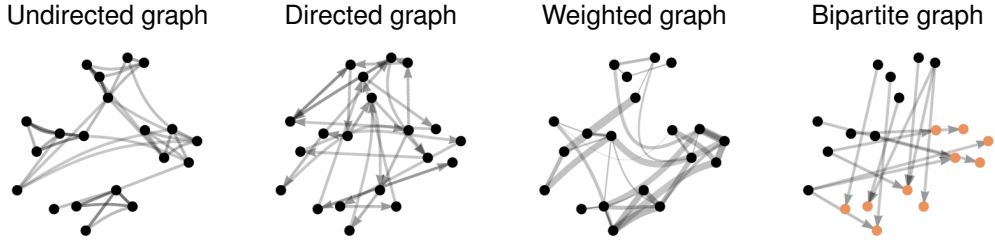
  \caption{Illustration of undirected, directed, weighted, and
    bipartite graphs.  An undirected graph consists of a set of nodes
    and a set of edges. In directed graphs, edges point from one node
    to another. Edges in a weighed graph have an associated value
    e.g. representing the strength of the relation. A bipartite graph
    represents a set of relations between two disjoint sets of
    nodes. Non-parametric Bayesian models can be formulated for all of
    these types of network structures.}
  \label{fig:NetworkTypes}
\end{figure*}

A complex network can be represented as a graph $G(V,E)$ with vertices (nodes) $V$ and edges (links) $E$ where an edge defines a connection between two of the vertices. In the following we denote the number of nodes in the graph by $N$ and the number of links by $L$. Graphs are often represented in terms of their corresponding adjacency matrix $X$ defined such that $x_{i,j}=1$ if there exists a link between node $i$ and $j$ and $x_{i,j}=0$ otherwise. Common types of graphs include undirected, directed, and bipartite graphs, and these can in turn be weighted such that each link has an associated strength (see Figure~\ref{fig:NetworkTypes}).  Complex networks are commonly stored in a sparse representation as an ``edge list''; a set of $L$ 3-tuples $(i,j,w)$ where $w$ is the weight of the link from node $i$ to node $j$. Using this representation, the storage requirements for a network grows linearly in the number of edges of the graph.

\subsection{Network characteristics}
\begin{figure} 
  \centering
  \begin{tabular}{m{10mm}b{5mm}b{5mm}b{5mm}b{24mm}}
    \toprule
    &
    \rotatebox{30}{\sffamily Degree mean} &
    \rotatebox{30}{\sffamily Degree std.} &
    \rotatebox{30}{\sffamily Clustering coef.} &
    \rotatebox{30}{\sffamily Char. path length}\\
    \midrule
    \multicolumn{5}{l}{\sffamily Erd\H{o}s-R\'{e}nyi graph}\\
    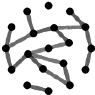 & 1.8 & 1.1 & 0.0 & 2.9\\
    \multicolumn{5}{l}{\sffamily Heavy tailed degree distribution}\\
    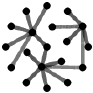 & 1.9 & \bfseries 1.9 & 0.0 & 3.1\\
    \multicolumn{5}{l}{\sffamily High clustering coefficient}\\
    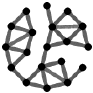 & 3.1 & 1.2 & \bfseries 0.5 & 3.4\\
    \multicolumn{5}{l}{\sffamily Long characteristic path length}\\
    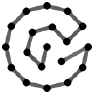 & 1.9 & 0.3 & 0.0 & \bfseries 6.3\\
    \bottomrule
  \end{tabular}
  \caption{Illustration of three important network characteristics:
    the degree distribution, clustering coefficient, and
    characteristic path length. The \emph{degree} of a vertex is the
    number of edges that links it to the rest of the network. The
    \emph{clustering coefficient}, defined as the average fraction of
    triangles relative to the total number of potential triangles
    given the vertex degree, quantifies the degree to which the
    vertices in a graph tend to cluster together.  The
    \emph{characteristic path length} is defined as the average
    shortest path between the vertices of the
    network.}\label{fig:NetworkCharacteristics}
\end{figure}

An important regimen in network science is to examine different characteristics or metrics computed from an observed network. The characteristics that have been examined include the distribution of the number of edges for each vertex (the degree distribution), the tendency of vertices to cluster together in tightly knit groups (the clustering coefficient), the average number of links required to move from one vertex to another (the characteristic path length), and many more (see Figure~\ref{fig:NetworkCharacteristics}, and for a detailed list of studied network characteristics see \cite{Rubinov2010}.)

To assess the importance of these characteristics they can be contrasted with the properties of some class of random graphs: To discover significant properties which cannot be explained by pure chance. The most simple class of random graphs used for comparison is the socalled Erd\H{o}s-R\'{e}nyi graphs in which pairs of nodes connect independently at random with a given connection probability $\phi$,
\begin{equation}
  x_{i,j}\sim \Bernoulli(\phi), \quad \phi\in[0;1].
\end{equation}

Amongst the findings is that many real networks exhibit ``scale free'' and ``small-world'' properties. A network is said to be scale free if its degree distribution follows a power law~\cite{Barabasi1999} in contrast to Erd\H{o}s-R\'{e}nyi random graphs which have a binomial degree distribution. The power law degree distribution indicates that many nodes have very few links whereas a few nodes (hubs) have a large number of links. A network is said to be small-world if it has local connectivity and global reach such that any node can be reached from any other node in a small number of steps along the edges. This associates with having a large clustering coefficient and small characteristic path length \cite{Watts1998} and suggests that generic organizing principles and growth mechanisms may give rise to the structure of many existing networks \cite{Watts1998,Barabasi1999,sporns2010networks,Eguiluz2005,borner2007network,leskovec2008dynamics}. Using analytic tools from network science, studies have demonstrated that many complex networks behave far from random \cite{Watts1998,Barabasi1999,sporns2010networks,Eguiluz2005}.

\subsection{Exponential random graphs}
To understand the processes that govern the formation of links in complex networks, statistical models consider some class of probability distributions over networks. A prominent and very rich, general class of models for networks is the exponential random graph family \cite{frank1986markov,robins2007recent,wasserman2005introduction}, also denoted the $p^*$ model. In the exponential random graph model the probability of an observed network takes the form of an exponential family distribution,
\begin{equation}
  p(X|\theta) = \frac{1}{\kappa(\theta)}\exp\left\{\theta^\top s(X)\right\},
\end{equation}
where $\theta$ is a vector of parameters, $s(X)$ is a vector of sufficient statistics, and $\kappa(\theta)$ is the normalizing constant that ensures that the distribution sums to unity. In general, the sufficient statistic can depend on three different types of quantities:
\begin{LaTeXdescription}
\item[Exogenous predictors:] In addition to the network, side information is often available which can aid in modeling the network structure. Including such observed covariates on the node or dyad level allows the analysis of networks and side information in a single model.
\item[Network statistics:] Statistics computed on the network itself, such as counts of different network motifs can be included. This could be quantities such as the number of edges, triangles, two-stars, etc. Since these terms depend on the graph, they introduce a self-dependency in the model, significantly complicating the inference procedure. There is virtually no limit to which terms could potentially be included, and how to choose a suitable set terms for a specific network domain is an open problem.
\item[Latent variables:] The network can be endowed with a latent structure that characterizes the network generating process. The latent variables could for example be continuous or categorical variables on the node level or a latent hierarchical structure.  The latent variables are most often jointly inferred with the model parameters. One reason for including latent variables is to aid in the understaning of the model: For example, if each network node is given a categorical latent variable, this corresponds to a clustering of the network nodes.
\end{LaTeXdescription}

The parameters in exponential random graphs are usually estimated using maximum likelihood which can be non-trivial since the normalizing constant usually can not be explicitly evaluated. While exponential random graph models are very flexible and work well for predicting links, they have the following important shortcomings:
\begin{LaTeXdescription}
\item[Model complexity:] It can be difficult to determine the suitable model complexity: Which network statistics to include, how many latent dimensions or categories to include etc. To address this issue different approaches have been taken, including imposing sparsity on the parameters and using model order selection tools such as BIC and AIC.
\item[Computational complexity:] In general, the computational complexity of inference in exponential random graph models grows with the size of the network, $\mathcal{O}(N^2)$, rather than with the number of edges, $\mathcal{O}(L)$, making exact large scale analysis infeasible. There are, however, certain special cases for which the complexity of inference scales linearly in the number of edges, which we will discuss further in the sequel.
\item[Inferential complexity:] When only exogenous predictors and latent variables are included in the model, inference is fairly straightforward; however, when \emph{network statistics} are included inference can be challenging, involving either heuristics such as pseudo likelihood estimation or complicated Markov chain Monte Carlo methods~\cite{robins2007recent,robins2007introduction}.
\end{LaTeXdescription}

\subsection{Non-parametric Bayesian network models}
In the following we present a number of recent network modeling approaches based on Bayesian non-parametrics which can all be seen as extensions or special cases of the exponential random graph model. In non-parametric modeling, the structure of the model is not fixed, and thus the model complexity can adapt as needed according to the complexity of the data. This forms a principled framework for addressing the first issue (model complexity) mentioned above. With respect to the second issue (computational complexity), it turns out that many of these non-parametric Bayesian models can be constructed such that their computational complexity is linear in the number of links, allowing these methods to scale to large networks. While it certainly is possible to include network statistics in non-parametric Bayesian network models, Bayesian non-parametrics does not address the third issue (inferential complextiy) which is an open area of research.

The focus of the remainder of this paper is twofold:~i)~To provide a comprehensible tutorial on the most simple non-parametric Bayesian network model: The infinite relational model~\cite{kemp2006learning,xu2006learning}. ~ii)~To give a brief overview of current advances in non-parametric Bayesian network models.

\begin{figure*} 
  \centering
  \begin{minipage}{1.0\linewidth}
    \textsf{\large Bayesian modeling}
    \label{sec:Bayes}
    \begin{multicols}{3}

      In traditional frequentist statistical modeling, probabilities
      describe relative frequencies of random variables in the limit
      of infinitely many trials. Model parameters are considered
      unknown but fixed quantities. A statistical model is
      characterized by a set of distributions,
      \begin{align}
        \label{eq:likelihood}
        p(X|\theta), & & \mathrm{(likelihood)}
      \end{align}
      where the unknown parameter $\theta$ takes values in parameter
      space $\Theta$. When considered as a function of $\theta$, the
      distribution $p(X|\theta)$ is known as the \emph{likelihood}. A
      non-parametric model is, contrary to what one might expect from
      its name, not a model without parameters, but a model which can
      not be parameterized by a finite dimensional parameter space. In
      other words, we can think of a non-parametric model as one
      having an infinite number of parameters---a notion that will be
      made explicit later.

      In Bayesian modeling, in addition to describing random
      variables, probabilities are used to describe inferences, i.e.,
      to quantify degree of belief about the parameters. Although
      parameters are still thought of as unknown, fixed quantities,
      they are modeled as random variables where the randomness
      reflects our lack of knowledge about them. To this end, they are
      assigned a socalled \emph{prior} probability distribution,
      \begin{align}
        \label{eq:prior}
        p(\theta), & & \mathrm{(prior)}
      \end{align}
      representing the degree of belief about the model parameters
      prior to observing any data. Often, it is convenient to specify
      the prior using some parameterized family of distributions. The
      parameters of the prior distribution are often referred to as
      \emph{hyper parameters} and can either be fixed or assigned
      \emph{hyper priors} which themselves might have hyper-hyper
      parameters, etc. A model defined in this manner is referred to
      as a hierarchical Bayesian model.

      Once the prior and the likelihood have been decided upon, the
      model is completely specified. Inference entails using the rules
      of probability to compute the conditional distribution of the
      parameters given the observations, also known as the
      \emph{posterior},
      \begin{align}
        \label{eq:posterior}
        p(\theta|x) &
        =\frac{p(x|\theta)p(\theta)}{\int p(x|\theta)p(\theta)\mathrm{d}\theta},\\
        \mathrm{posterior} &
        =\frac{\mathrm{likelihood}\times\mathrm{prior}}{\mathrm{evidence}}.
      \end{align}
      Thus, we are not merely interested in a single parameter
      estimate, but aim at estimating a distribution over parameters
      quantifying our state of knowledge about the parameters after
      observing the data.

      Often, only a subset of the parameters is of intereset---the
      others are simply used as a means to specifying a reasonable
      probabilistic model, but are not of interest themselves. Such
      parameters are often referred to as \emph{nuisance}
      parameters. Assume for instance the parameters
      $\theta=\{\iota,\nu\}$ can be divided into interesting~($\iota$)
      and nuisance~($\nu$) parameters. In that case, we compute the
      posterior distribution of the parameters of interest,
      \begin{equation}
        p(\iota|x) = \int p(\theta|x) \dd \nu,
      \end{equation}
      which can be found by marginalizing (integrating over) the
      nuisance parameters.

      Although conceptually simple, inference might be computationally
      unwieldy because of high dimensional and analytically
      intractable integrals (or summations, in the case of discrete
      parameters). In practice one must therefore use some method of
      approximation, which we will discuss later.

      Bayesian data modeling can be divided into three
      tasks~\cite{Gelman2004}:
      \paragraph{Joint distribution:} The first step involves
      formulating the probabilistic model, i.e. a joint distribution
      over data and parameters, by specifying the likelihood and
      priors.
      \paragraph{Inference:} Next, the posterior distribution of the
      parameters is inferred, often using some method of numerical
      approximation such as Monte Carlo sampling.
      \paragraph{Checking implications:} Finally we check how well the
      model describe the data and evaluate the implications of the
      posterior distribution by computing quantities of interest and
      making decisions.  \vspace{2mm}

      In this paper we go through the details of these three steps in
      the context of the infinite relational
      model~\cite{kemp2006learning,xu2006learning}.
    \end{multicols}
  \end{minipage}
  \caption{A brief introduction to Bayesian modeling introducing the
    concepts needed in this paper.}
  \label{fig:Bayes}
\end{figure*}

\section{Tutorial on the Infinite relational model}
In the following we give a tutorial introduction to the infinite relational model~\cite{kemp2006learning,xu2006learning} which is perhaps the most simple non-parametric Bayesian network model. We will derive the necessary Bayesian non-parametric machinery from first principles by taking limits of a parametric Bayesian model. Understanding the details of involved in deriving this simple model later serves as a foundation for understanding other more complicated non-parametric constructions. Further, we go though the details involved in inference by Markov chain Monte Carlo, and show how a Gibbs sampler can be implemented in a few lines of computer code. Finally, we demonstrate the model on three network datasets and compare with other models from the exponential random graph model family.

\subsection{The infinite relational model}
The infinite relational model is a latent variable model where each node is assigned to a category, corresponding to a clustering of the network nodes. The number of clusters is learned from data as part of the statistical inference. As a starting point, we introduce a Bayesian parametric version of the model, which we later extend to the non-parametric setting.  For readers unaccustomed with Bayesian modeling, we provide a short introduction, see Figure~\ref{fig:Bayes}.

\subsubsection{A parametric Bayesian stochastic blockmodel}
\label{sec:BayesStochasticBM}
A simple and very powerful approach to modeling structure in a complex network is to use a \emph{mixture model}, leading to a Bayesian version of the socalled \emph{stochastic blockmodel} \cite{Nowicki2001}. In a mixture model, the observations are assumed to be distributed according to a mixture of $K$ components belonging to some parametric family. Conditioned on knowing which mixture components generated each datum, the observations are assumed independent. In a mixture model for network data, each node belongs to a single mixture component, and since each edge is associated with two nodes, its likelihood will depend on two components. Thus, the likelihood of the network will take the following form,
\begin{equation}
  \label{eq:mixture}
  p(X|\theta) = \prod_{(i,j)} p(x_{i,j}|z_i,z_j,\phi)
\end{equation}
where the product ranges over all node pairs, and the parameters are given by $\theta=\big\{\{z_i\}_{i=1}^{N},\phi\big\}$ where $z_i$ indicates which mixture component the $i$th node belongs to and $\phi$ denotes any further parameters. In the most simple setting, each term in the likelihood could be a Bernoulli distribution (a biased coin flip),
\begin{align}
  p(x_{i,j}|z_i,z_j,\phi) &= \Bernoulli(\phi_{z_i,z_j})\\
  &= (\phi_{z_i,z_j})^{x_{i,j}}(1-\phi_{z_i,z_j})^{1-x_{i,j}},
\end{align}
such that $\phi_{k,\ell}$ denotes the probability of an edge between two nodes in group $k$ and $\ell$. To finish the specification of the model, we must define prior distributions for the mixture component indicators $z$ as well as the link probabilities $\phi$. Starting with $\phi$, a natural choice would be independent $\Beta$ distributions for each pair of components,
\begin{align}
  p(\phi_{k,\ell}) &= \Beta(a,b) \\
  &= \frac{1}{\B(a,b)}(\phi_{k,\ell})^{a-1}(1-\phi_{k,\ell})^{b-1},
\end{align}
where the parameters for example can be set to $a=b=1$ to yield a uniform distribution. A natural choice for $z$ would be a $K$-dimensional categorical distribution,
\begin{equation}
  \label{eq:zprior}
  p(z_i=k|\pi) = \pi_k
\end{equation}
parameterized by $\pi=\{\pi_k\}_{k=1}^{K}$ where $\sum_{k=1}^{K}\pi_k=1$. How, then, should $\pi$ be chosen? We could for example set each of these parameters to a fixed value, e.g. $\pi_k=\tfrac{1}{K}$, but this would be a strong prior assumption specifying that the mixture components have the same number of members on average. A more flexible option would be to define a hierarchical prior, where $\pi$ is generated from a Dirichlet distribution,
\begin{align}
  p(\pi) &= \Dirichlet(\alpha)\\
  &= \frac{1}{\B(\alpha)}\prod_{k=1}^{K}\pi_k^{\alpha_k-1}.
\end{align}
where $\B(\alpha)$ is the multinomial beta function, which can be expressed using the gamma function,
\begin{equation}
  \B(\alpha)=\frac{\prod_{k=1}^{K}\Gamma(\alpha_k)}{\Gamma(\sum_{k=1}^{K}\alpha_k)}.
\end{equation}
Since each component a priori is equally likely, we select the concentration parameters to be equal to each other, $\alpha_1=\dots=\alpha_K=\tfrac{A}{K}$ such that the scale of the distribution is $\sum_{k=1}^{K}\alpha_k=A$. This results in a joint prior over $z$ and $\pi$ given by
\begin{align}
  p(z,\pi) &= \left[\prod_{i=1}^{N}p(z_i|\pi)\right]\times p(\pi|\alpha)\\
  &= \frac{1}{\B(\alpha)}\prod_{k=1}^{K}\pi_k^{n_k+\alpha_k-1},
\end{align}
where $n_k$ denotes the number of $z_i$'s with the value $k$.

\subsubsection{Nuisance parameters}
As we are not particularly interested in the mixture component probabilities $\pi$ (they can be considered nuisance parameters) we can compute the effective prior over $z$ by marginalizing over $\pi$ which has a closed form expression due to the conjugacy between the Dirichlet and Categorical distributions (i.e., the posterior distribution of $\pi$ has the same functional form as the prior),
\begin{align}
  p(z) &= \int p(z,\pi) \dd\pi =
  \frac{\B(\alpha+n)}{\B(\alpha)} \\
  &= \frac{\Gamma(A)}{\Gamma(A+N)}
  \prod_{k=1}^{K} \frac{\Gamma(\alpha_k+n_k)}{\Gamma(\alpha_k)}.
  \label{eq:p(z)}
\end{align}
This resulting effective prior distribution is known as a multivariate P\'{o}lya distribution.

Furthermore, the link probabilities $\phi$ can also be considered nuisance parameters, and can also be marginalized analytically due to the conjugacy between the Beta and Bernoulli distributions,
\begin{align}
  p(X|z) &= \int p(X|z,\phi)p(\phi)\dd \phi\\
  &= \prod_{(k,\ell)}\frac{\B(m_{k,\ell}+a,\bar m_{k,\ell}+b)}{\B(a,b)},
  \label{eq:p(x|z)}
\end{align}
where the product ranges over all pairs of components and $m_{k,\ell}$ and $\bar m_{k,\ell}$ denote the number of links and non-links between nodes in component $k$ and $\ell$ respectively.

\subsubsection{An infinite number of components}
\label{sec:NPBayes}
In the previous section we specified a \emph{parametric} Bayesian mixture model for complex networks. In the following we move to the non-parametric setting in which the number of mixture components is allowed to be countably infinite. First, consider what happens when the number of components is much larger than the number of nodes in the graph. In that situation, many of the components will not have any nodes assigned to them; in fact, no more than $N$ components can be non-empty, corresponding to the worst case situation where each node has a component of its own. To handle the situation with an infinite number of components, we can not explicitly represent the components but, as we will show in the following, we need only an explicit representation of the finite number of non-empty components.

As we defined the model so far we have introduced $K$ \emph{labelled} mixture components. This means that if we for example have $N=5$ nodes and $K=4$ components, we assign a separate probability to, say, the configurations $\{1,2,1,4,2\}$ and $\{3,4,3,2,4\}$ even though they correspond to the same clustering of the network nodes. A better choice is to specify the probability distribution directly over the \emph{equivalence class} of partitions of the network nodes. Since we have $K$ labels in total to choose from, there are $K$ possible labellings for the first component, $K-1$ for the second, etc. resulting in a total of
\begin{equation}
  \frac{K!}{(K-\bar K)!}
\end{equation}
labellings corresponding to the same partitioning, where $\bar K$ is the number of non-empty components. Thus, defining a parameter $\bar z$ that holds the partitioning of the network nodes, we have
\begin{align}
  p(\bar z) = \frac{K!}{(K-\bar K)!}
  \frac{\Gamma(A)}{\Gamma(A+N)} \prod_{k=1}^{K}
  \frac{\Gamma(\alpha_k+n_k)}{\Gamma(\alpha_k)}.
  \label{eq:CRPPrior}
\end{align}
Since $\bar z$ represents partitions rather than labels it can be finitely represented.  We can now simply let the number of components go to infinity by computing the limit of the prior distribution for $\bar z$,
\begin{align}
  \lim_{K\rightarrow \infty} p(\bar z) =
  \frac{\Gamma(A)A^{\bar K}}{\Gamma(A+N)} \prod_{k=1}^{\bar K}
  \Gamma(n_k).
\end{align}
The details involved in computing this limit can be found in \cite{Green2001} and \cite{Neal1992}. The limiting stochastic process is known as a \emph{Chinese restaurant process} (CRP) \cite{aldous1985} (for an introduction to the CRP, see \cite{Gershman2011}). Compactly, we may write
\begin{align}
  \bar z & \sim \CRP(A).
\end{align}

\begin{figure} 
  \centering
  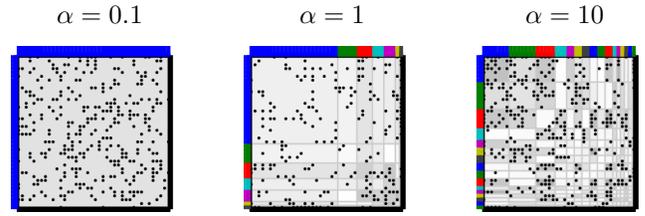
  \caption{Example of graphs generated according to the \emph{infinite
      relational model} for different choices of the parameter
    $\alpha$ of the Chinese Restaurant Process.}
  \label{fig:fig4}
\end{figure}

\subsubsection{Summary of the generative model}
In summary, the generative process for the infinite relational model can be expressed as follows:
\begin{align}
  \bar z & \sim \CRP(A),\\
  \phi_{k,\ell}  & \sim \Beta(a,b),\\
  x_{i,j} & \sim \Bernoulli(\phi_{z_i,z_j}).\label{eq:irm_likelihhod_final}
\end{align}
The network nodes are partitioned according to a Chinese restaurant process; a probability of linking between each pair of node clusters is simulated from a Beta distribution; and each link in the network is generated according to a Bernoulli distribution depending on which clusters the pair of nodes belong to.

Identically, in the notation of exponential random graph models the likelihood in Eq.~(\ref{eq:irm_likelihhod_final}) can be expressed as
\begin{equation}
  \label{eq:IRM_as_ERGM}
  p(X|z,\phi) = \frac{1}{\kappa(z,\phi)}\exp\left[\theta(\phi)^\top s(X,z)\right]
\end{equation}
where the sufficient statistics are the counts of links between each pair of clusters, $s(X,z) = \{m_{k,\ell}\}$, and the natural parameter is the log odds of links between each pair of clusters, $\theta(\phi)=\left\{\log\tfrac{\phi_{k,\ell}}{1-\phi_{k,\ell}}\right\}$.

\subsection{Inference}
Having specified the model in terms of the joint distribution, the next step is to examine the posterior distribution which is given as
\begin{align}
  p(\bar z|X) = \frac{p(X|\bar z)p(\bar z)}{\displaystyle\sum_{\bar z}
    p(X|\bar z)p(\bar z)}.
\end{align}
Here, the numerator is easy to compute as the product of Eq.~(\ref{eq:p(x|z)}) and (\ref{eq:CRPPrior}); however, the denominator is difficult to handle as it involves an elaborate summation over all possible node partitionings. Consequently, some form of approximate inference is needed.

There are two major paradigms in approximate inference: Variational and Monte Carlo inference.  The idea in variational inference is to approximate the posterior distribution with a simple, tractable distribution which is fitted to the posterior by minimizing some criterion such as the information divergence~\cite{Blei2006}.

In Monte Carlo approximation the idea is to generate a number of random samples from the posterior distribution and approximate intractable integrals and summations by empirical averages based on the samples.

In the following, we focus on Monte Carlo inference. In particular we review the Gibbs sampler for the infinite relational model.

\subsubsection{Gibbs sampling}
In Gibbs sampling the variables are iteratively sampled from their conditional distribution, and repeating this process the samples will eventually approximate the posterior distribution. We iteratively sample the partition assignments, $\bar z_n$, from their conditional distribution,
\begin{align}
  p(\bar z_n = k|\bar z^{\setminus n},X),
\end{align}
where $\bar z^{\setminus n}$ denotes all partition assignments except $\bar z_n$. An expression for this conditional distribution can be found by considering which terms in the likelihood and prior will change when node $n$ is assigned to a different partition. For the prior in Eq.~(\ref{eq:CRPPrior}) we have
\begin{align}
  p(\bar z_n = k|\bar z^{\setminus n}) \propto
  \left\{\begin{array}{ll}
      n^-_k & k \text{ is an existing partition,} \\
      A     & k \text{ is a new partition,}
    \end{array}
  \right.
  \label{eq:GibbsPrior}
\end{align}
where $n^-_k$ is the number of nodes associated with component $k$ not counting node $n$. Adding node $n$ to an existing component increases the argument of the corresponding Gamma function by one, effectively multiplying the prior by $n^-_k$, whereas adding the node to a new cluster increases $\bar K$ by one, effectively multiplying the prior by $A$. For the likelihood in Eq.~(\ref{eq:p(x|z)}), adding node $n$ to partition $k$ effectively multiplies the likelihood by
\begin{align}
  \prod_\ell \frac{\B\left(m^{\setminus
        n}_{k,\ell}\!+\!r_{n,\ell}\!+\!a,\ \bar m^{\setminus
        n}_{k,\ell}\!+\!n_\ell\!-\!r_{n,\ell}\!+\!b\right)}
  {\B\left(m^{\setminus n}_{k,\ell}\!+\!a,\ \bar m^{\setminus
        n}_{k,\ell}\!+\!b\right)},
  \label{eq:GibbsLikelihood}
\end{align}
where $m^{\setminus n}_{k,\ell}$ and $\bar m^{\setminus n}_{k,\ell}$ denote the number of links and non-links between nodes in component $k$ and $\ell$, not counting any links from node $n$, and $r_{n,\ell}$ is the number of links from node $n$ to any nodes in component $\ell$.

In order to perform Gibbs sampling we can now simply consider each node in turn; for each partition (including a new, empty partition) compute the product of Eq.~(\ref{eq:GibbsPrior}) and (\ref{eq:GibbsLikelihood}); normalize to yield a categorical distribution over partitions; and sample a new $\bar z_n$ according to this distribution. The final algorithm is summarized in Fig.~\ref{fig:GibbsAlgorithm}. The result after running the Gibbs sampler for $2T$ iterations is a set of samples of $\bar z$, where usually the first half is discarded for burn in. This yields a final ensemble $\{\bar z^{(t)}: t\in 1,\dots,T\}$ approximately sampled from the posterior.

\begin{figure*} 
  \centering
\begin{lstlisting}
function Z = irm(X,T,a,b,A)
N = size(X,1); z = true(N,1); Z = cell(T,1);                  % Initialization
for t = 1:T                                                   % For each Gibbs sweep    
  for n = 1:N                                                 % For each node in the graph        
    nn = [1:n-1 n+1:N];                                       % All indices except n        
    K = size(z,2);                                            % No. of components
    m = sum(z(nn,:))'; M = repmat(m,1,K);                     % No. of nodes in each component
    M1 = z(nn,:)'*X(nn,nn)*z(nn,:)- ...                       % No. of links between components
        diag(sum(X(nn,nn)*z(nn,:).*z(nn,:))/2);
    M0 = m*m'-diag(m.*(m+1)/2) - M1;                          % No. of non-links between components
    r = z(nn,:)'*X(nn,n); R = repmat(r,1,K);                  % No. of links from node n    
    logP = sum([betaln(M1+R+a,M0+M-R+b)-betaln(M1+a,M0+b) ... % Log probability of n belonging
      betaln(r+a,m-r+b)-betaln(a,b)],1)' + log([m; A]);       %   to existing or new component
    P = exp(logP-max(logP));                                  % Convert from log probability
    i = find(rand<cumsum(P)/sum(P),1);                        % Random component according to P
    z(n,:) = false;  z(n,i) = true;                           % Update assignment        
    z(:,sum(z)==0) = [];                                      % Remove any empty components                
  end
  Z{t} = z;                                                   % Save result 
end
\end{lstlisting}
  \caption{\matlab\ code implementing the infinite relational model. $X$
    is the symmetric adjacency matrix, $T$ is the number of Gibbs
    sweeps, and $a$, $b$, and $A$ are the hyperparameters. The code
    illustrates the computations involved in the Gibbs sampler, but is
    not efficient since it recomputes all the needed link counts in
    each iteration.}
  \label{fig:GibbsAlgorithm}
\end{figure*}

\subsubsection{Computational complexity}
In the algorithm outlined in Figure~\ref{fig:GibbsAlgorithm} it can be observed that there are two loops: One over the $T$ simulated samples and one over the $N$ nodes in the network. In each run of the inner loop, a node is assigned to a cluster by the Gibbs sampler. In the following we consider the number of clusters $K$ a constant (although of course it will vary depending on the network data), and examine how the computational complexity of the algorithm depends on the number of nodes and edges in the network.

In the code in Figure~\ref{fig:GibbsAlgorithm} the variables \texttt{M0}, \texttt{M1} and \texttt{m}, which hold the counts of nonlinks, links, and nodes, are re-computed in each iteration. In a more sensible implementation, these quantities would be precomputed and efficiently updated during the Gibbs sampling.

\begin{figure} 
  \centering
  \footnotesize
  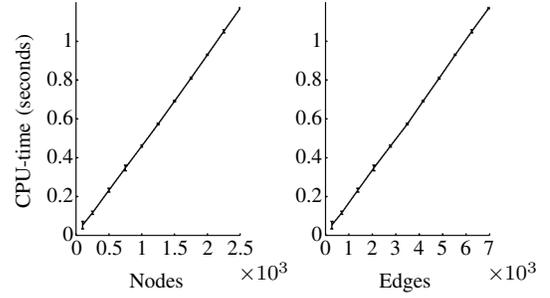
  \caption{Experiment demonstrating that the computational complexity
    grows linear in both the number of nodes $N$ and edges $L$ for the
    IRM model. The graphs used in the experiments are generated with
    $K=5$ communities of equal size and $\phi=\phi_c/N$ where $\phi_c$
    is kept constant in the experiments ensuring that the number of
    edges $L$ grows linearly with the number of nodes $N$ in the
    generated networks. The Gibbs sampler used in the experiment was
    implemented to pre-compute \texttt{M0}, \texttt{M1} and \texttt{m}
    resulting in a computational complexity of $\mathrm{O}(L)$ for
    each iteration of the sampler. Given are the mean CPU-times in
    seconds for the sampler and standard deviation across $T=10$
    iterations when varying the number of nodes ($N$) and edges ($L$)
    in the generated graphs.}
  \label{fig:fig6}
\end{figure}

Evaluating the probability of assigning a node to each cluster then requires the computation of the vector \texttt{r} which holds the count of links from node $n$ to each of the clusters. The time complexity of this computation is on the order of the node degree. Looping over the nodes gives a total time complexity of $\mathrm{O}(L)$ where $L$ is the number of edges in the graph.  To calculate the probabilities of assigning the nodes to the clusters for all $N$ Gibbs samples requires $2K^2N$ evaluations of the (logarithm of the) Beta function so the time complexity of this computation is $\mathrm{O}(N)$. As a result, since in general $L>N$, the total computational complexity of the Gibbs sampler for the IRM model is $\mathrm{O}(L)$. Figure~\ref{fig:fig6} demonstrates that this linear scaling is observed in practice when analyzing networks of varying numbers of nodes and edges.

 For comparison, Monte Carlo maximum likelihood inference in exponential random graph model based on endogenous network statistics requires the simulation of random networks from the ERGM distribution, which is inherently an $\mathrm{O}(N^2)$ operation. We should note though, that in practice we would not expect the time complexity of the IRM to scale linearly in the number of edges, since the number of clusters most likely would increase with the size of the network and since the number of required iterations of the Gibbs sampler might also go up.

\subsection{Checking model fit}
Once an approximation of the posterior distribution has been obtained, we wish to check the implications of the model. This can include computing the posterior distribution of important quantities of interest, evaluating how well the model fits the data, and making predictions about unobserved data.

\subsubsection{Computing posterior quantities}
Say we are interested in some function $f(\bar z)$ that depends on the model. We can now compute the posterior distribution of this quantity,
\begin{align}
  p\left(f(\bar z)\right) &= \sum_{\bar z^\prime} \delta(f(\bar z)=f(\bar z^\prime)) p(\bar z^\prime|X)\\
  &\approx \frac{1}{T}\sum_{t=1}^T \delta(f(\bar z)=f(\bar z^{(t)})),
\end{align}
approximated by an empirical average over the posterior samples. For example, the approach can be used to compute the posterior distribution over the number of components in the mixture model or other quantities of interest.

\subsubsection{Link prediction}
Missing data is easily handled in the Bayesian framework, simply by leaving out the terms in the likelihood corresponding to unobserved links. If we observe only a part of the network and are interested in predicting the presence or absence of an unobserved link between two nodes, we can simply compute the posterior predictive distribution of the missing link,
\begin{align}
  p(x_{i,j}|X) &= \sum_{\bar z} p(x_{i,j}|\bar z,X) p(\bar z|X) \\
  & \approx  \frac{1}{T}\sum_{t=1}^T p(x_{i,j}|\bar z^{(t)},X).
\end{align}
Here $X$ denotes the observed part of the network, and $\bar z^{(t)}$ is simulated from the posterior distribution where only the observed part of the network is conditioned on. Inserting $p(x_{i,j}|\bar z,X) = \int p(x_{i,j}|\theta,\bar z)p(\theta|\bar z,X)d\theta$ yields
\begin{equation}
  p(x_{ij}|X)\approx \Bernoulli(\rho_{i,j}),
  \label{eq:LinkPredictionBernoulli}
\end{equation}
where
\begin{equation}
  \rho_{i,j} = \frac{1}{T}\sum_t\frac{m_{z^{(t)}_i,z^{(t)}_j}+a}
    {m_{z^{(t)}_i,z^{(t)}_j}+\bar{m}_{z^{(t)}_i,z^{(t)}_j}+a+b}
\end{equation}
Predicting missing links can be used to compare different models: A number of links can be excluded when fitting the models which can then be compared by assessing their ability to predict the held-out links. Since the links in a network are highly correlated and because many networks exhibit a highly imbalanced distribution of links and nonlinks, care must be taken in choosing a hold out test set in an appropriate way. If the test set is chosen to balance the number of links and nonlinks, its distribution will not correspond to the full network which makes the absolute link prediction results difficult to interpret. Thus, although indicative of a model's predictive performance, this approach is perhaps best suited for the relative comparison of different models. If, on the other hand, several examples of full networks are available, a whole network can used as test data making the absolute link prediction results directly interpretable.

\subsubsection{Posterior predictive checking}
Finally, we might be interested in examining how well our model describes the data to assess if the model is appropriate for the data at hand or if a more suitable model should be constructed. A principled approach to achieving this is posterior predictive checking. First, an ensemble of replicated networks from the posterior predictive distribution is generated from,
\begin{align}
  p(X^{\mathrm{rep}}|X) = \sum_{\bar z}p(X^{\mathrm{rep}}|\bar z,X) p(\bar z|X),
\end{align}
which as before can be approximated using samples of $\bar z$ simulated from the posterior using Eq.~(\ref{eq:LinkPredictionBernoulli}). Now, the idea is to compare characteristics of the observed network, such as the degree distribution, clustering coefficient, and characteristic path length, with the posterior predictive distribution of these properties, approximated by the empirical distribution over the ensemble of replicated networks. If the model fits well, the observed characteristic of the network should be quite likely under the posterior predictive distribution, whereas a large discrepancy indicates model mismatch. Posterior predictive checking is useful for model \emph{critisism}, i.e., for exploring lack of fit as opposed to testing whether the model is correct. Discovering network characteristics for which the model does not fit the data well can inspire to the development of more sophisticated models; however, even a simple model which does not fit the data in all respects can be useful.

\subsection{Directed, weighted, bipartite, and multiple networks}
The infinite relational model readily extends to other types of graphs including directed, weighted and bipartite networks as well as multiple networks on the same set of nodes. These extensions can be arrived at by modifying the model parametrization and the observational model (the likelihood function) as well as making appropriate changes to the priors. The process of formulating the joint distribution and deriving a Markov chain Monte Carlo procedure for inference closely follows the steps we have taken for the basic infinite relational model described in the previous sections. The extensions described below can also be combined, for example to model a set of directed, bipartite networks with edge weights.

\subsubsection{Directed networks}
In a directed network, the links have an associated direction, so that they point from one node to another. A directed network can be represented by an asymmetric adjacency matrix, and the directionality of links between groups can be modelled through the parameter $\phi$ by the existence of asymmetric interactions between the groups such that $\phi_{k,\ell}\neq \phi_{\ell,k}$. This double the number of link probability parameters $\phi$. The rest of the model is unaffected, except for the likelihood which must now be evaluated not for each \emph{pair} of nodes but for each \emph{ordered pair} of nodes. This extension of the infinite relational model assigns different probabilities to links in each direction between each pair of clusters, but has only a single parameter for the link probability within each cluster---thus, directionality is not modelled within clusters.

\subsubsection{Bipartite networks}
A bipartite network is defined as a set of links between two disjoint sets of nodes, possibly with different cardinality. The adjacency matrix for a bipartite network can thus be non-square. We can then use two independent Chinese restaurant processes to model the clustering of the two sets of nodes,
\begin{equation}
  \bar z \sim \CRP(A_z),\quad \bar w \sim \CRP(A_w)
\end{equation}
and change the likelihood to (cf. Eq.~(\ref{eq:irm_likelihhod_final})
\begin{align}
  x_{i,j} \sim \Bernoulli(\phi_{z_i,w_j}).
\end{align}
This latter parameterization is also useful for the modeling of directed networks when the groupings of the nodes may be different for the rows and columns of the adjacency matrix.

\subsubsection{Weighted networks}
In a weighted network, each edge has a (scalar) weight associated with it. Depending on the type of weights, the Bernoulli likelihood can be changed to some other suitable distribution: For example, if the weights are positive integers~\cite{MMMNS_NECO2012}, a Poisson distribution could be employed,
\begin{align}
  x_{i,j} \sim \Poisson(\lambda_{z_i,z_j}),
\end{align}
where $\lambda$ is the rate parameter for the edge weights, playing the role of $\phi$ in the Bernoulli model, c.f. Eq.~(\ref{eq:irm_likelihhod_final}). As a prior over $\lambda$, the typical choice is a Gamma distribution, replacing the Beta priors for $\phi$. If the weights are real numbers~\cite{Tue_MLSP2012} an observational model based on a Normal distribution might be appropriate,
\begin{align}
  x_{i,j}\sim \Normal(\mu_{z_i,z_j},\sigma_{z_i,z_j}^2).
\end{align}
Here, we have two sets of parameters, $\mu$ and $\sigma^2$, denoting the means and variances of the edge weights between nodes in groups $i$ and $j$. Again, appropriate priors for $\mu$ and $\sigma^2$ should be selected.

\subsubsection{Multiple networks}
Sometimes the data consists of multiple observations of networks on the same set of nodes (see \cite{kemp2006learning,miller2009nonparametric,MMMNS_NECO2012,Andersen2012}). The only required change to the model is that the likelihood should be evaluated as the product of the likelihoods for each observed network.  It can then either be assumed that the clustering structure as well as the link probabilities are equal across the multiple networks, that the clustering structure is shared but the link probabilities only shared according to an additional clustering of the multiple graphs, or that the clustering is shared but each network has an individual set of link probabilities,$\phi$. When the link probabilities are analytically marginalized this leads to three different expressions for the marginal likelihood.

\subsection{Experimental evaluation}
In the following we conduct a series of experimental evaluations with the infinite relational model, highlighting some of its properties and comparing it with other models.

\subsubsection{Analysis of three example networks}
To demonstrate the non-parametric Bayesian modeling framework in
practise, we analyzed three real networks:
\begin{LaTeXdescription}
\item[Zachary's Karate Club:] Zachary's Karate club is an undirected unweighted network of friendships between 34 members of a karate club at a US university in the 1970s \cite{zachary1977information}. A total of 74 undirected links between the members of the Karate club are observed. In the analysis the standard IRM model was used.
\item[Connectome of Caenorhabditis Elegans:] The only complete connectome currently recorded of an organism is the directed integer weighted network of the 8,799 connections between the 302 neurons of the Caenorhabditis Elegans. The network has been compiled in \cite{Watts1998}. In the analysis the weighted IRM model with a Poisson likelihood and Gamma priors was used.
\item[Drugs and side effects:] The drugs and side effects network is a bipartite network on marketed medicines and their recorded adverse drug reactions extracted from public documents and package inserts. The network currently consists of 996 drugs and 4,199 side effects with 100,049 unweighted links between drugs and side effects \cite{kuhn2010side}. In the analysis the bipartite IRM model was used.
\end{LaTeXdescription}
These three networks in turn represent three important complex network application domains within social science, neuroscience and bio-informatics.

The parameters of the models were inferred by Markov chain Monte Carlo sampling such that 250 iterations were used as burn in for the sampler and 250 iterations for drawing samples from the posterior. To improve mixing the data was analyzed based on 5 randomly initialized runs. In addition to a Gibbs sampler (as described in Fig.~\ref{fig:GibbsAlgorithm}) the socalled split-merge sampler described in~\cite{kemp2006learning,Jain2004} was also employed. The hyper-parameters for the Beta distribution we set to $a=b=1$. The posterior distribution of the number of components was computed. For assessing model fit by prediction of missing links, excluded 10\% of links and an equivalent number of non-links in the analysis of the Zachary's Karate Club data and 5\% of links and an equivalent number of non-links in the two larger Connectome and Drug-side effects networks. For posterior predictive checking, we stored every 25th posterior sample and generated 20 replicated networks for each sample for each of the five random initializations. From the ensemble of these networks the distribution over the network characteristics; \emph{degree mean, degree standard deviation, characteristic path length and clustering coefficient} were calculated and compared to the true values of these quantities for the actual network.

\begin{figure*} 
  \centering
  \scriptsize
  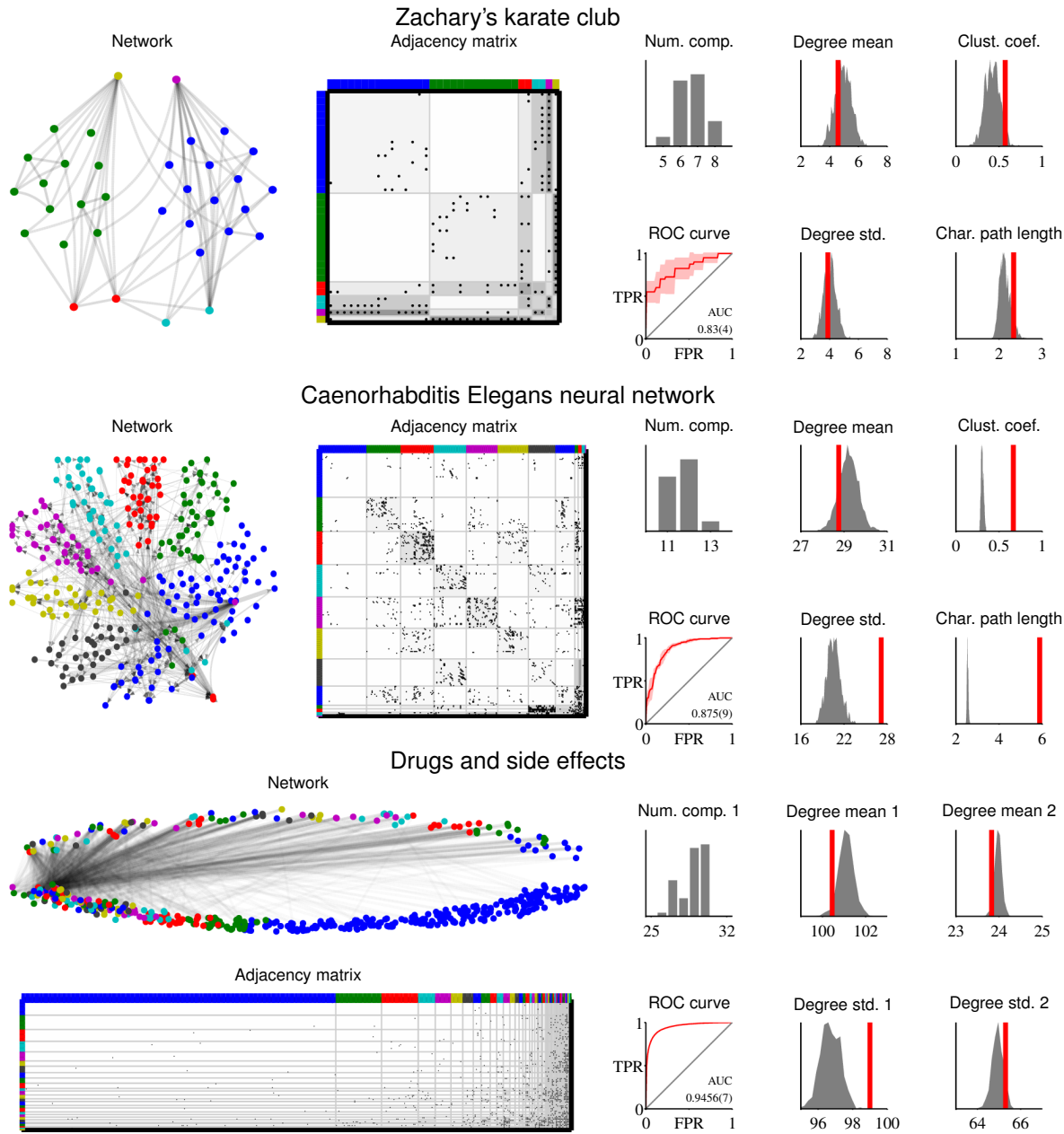
  \caption{Infinite relational model analysis of three networks:
    Social relations in Zachary's karate club, neural network of
    Caenorhabditis Elegans, and relations between drugs and side
    effects. Networks are shown as graphs (20 pct. of links shown for
    C. Elegans and 10 pct. shown for Drugs and side effects) as well
    as adjacency matrix. Posterior distribution of the number of
    components as well as ROC curve indicating performance on
    predicting missing links is shown (shaded regions indicate two
    times the standard deviation on the mean across the separate
    runs).  Posterior predictive distribution of node degree (mean and
    standard deviation), clustering coefficient, and characteristic
    path length is shown with vertical lines indicating values for the
    observed networks.}
  \label{fig:ZacharyCelegansDrugs}
\end{figure*}

The results of the modeling is given in Fig.~\ref{fig:ZacharyCelegansDrugs}. The figure illustrates the network as well as a permutation of the networks adjacency matrix. The nodes are color coded according to the partition given by the sample with highest posterior likelihood across the five random initializations. From the permuted adjacency matrices it can be seen that the nodes of the networks have been grouped into clusters that share similar patterns of interactions, defining regions of network homogeneities. These blocks are color coded according to the expected value of the corresponding group interactions using a logarithmic gray scale. On the right, the posterior distribution of the number of components is shown as well as the models performance in predicting held-out links. The link prediction performance is quantified by the area under curve (AUC) of the receiver operator characteristic (ROC)~\cite{miller2009nonparametric}. In addition, the results of the posterior predictive checking of the models ability to account for the mean and standard deviation of the degree distribution, characteristic path length and clustering coefficient are given. Since the degree is explicitly modelled in the IRM model, this posterior predictive check serves only as a sanity check: The IRM should by definition get this right except for a small bias due to the prior.

From these results it can be seen that the IRM model accounts well for all the considered characteristics in the Zachary Karate Club network but that it poorly accounts for the degree standard deviation, the clustering coefficient, and the characteristic path length of the connectome of C. Elegans. As expected, the average degree falls within the lower tails of the simulated distributions for all the estimated models, but they underestimate the standard deviation of the node degree of both the connectome and drugs and side effect networks. This highlights a deficiency of the IRM model, namely that it does not explicitly model the degree distribution.

While the IRM is adept in identifying blocks of homogeneous network regions with the $\phi$ (nuisance) parameter specifying the density of each of these blocks, it does not explicitly account for microscale properties such as triangles and node degree. Hence, the clustering coefficient, characteristic path length and the standard deviation of the degree distribution is not well accounted for by the model as is evident in the posterior predictive checks. Despite these limitations, the infinite relational model does well account for mesoscale structure in the networks as quantified by its ability to predict links: For all the three networks the infinite relational model is able to predict links significantly better than random guessing.

Apart from being able to predict links the IRM model has made the structure of the networks substantially more comprehensible by reducing the complex network of pairwise interactions to a much smaller number of groups (defined by $\bar z$) and their interactions (defined by $\phi$). The IRM can therefore be considered an efficient framework for compressing a large complex network to a smaller network constituting consistent patterns of interactions between groups of nodes which can substantially facilitate in the understanding of mesoscale network patterns. For example, the analysis of the Zachary's Karate Club network, the infinite relational model reveals six groups of club members including two large groups and two singletons (actually the posterior has support for five to eight groups, so these other configurations should also be considered in the interpretation of the results). It is known from the literature~\cite{zachary1977information} that the karate club later split into two fractions, corresponding to the two large groups, led by the president and the instructor which are the two singletons.

\subsubsection{Comparison with other models}
Next, we compare the IRM model to several other methods on a set of social networks derived from a study of intra-organizational relations:
\begin{LaTeXdescription}
\item[Intra-organizational relations:] This set of undirected networks \cite{Cross_Parker_2004} consists of two types of relations defined on the same set of nodes corresponding to employees in a consultancy company. Links in the first network signifies employees who iteract whereas links in the second network signifies that either of the employees thinks that the other has expertise in an area important to her. The networks were generated by thresholding and symmetrizing the original directed weighted networks~\cite{Cross_Parker_2004}. The two networks are highly correllated since employees would be expected to interact frequently with colleagues with important expertise.
\end{LaTeXdescription}

\begin{figure} 
  \centering
  \scriptsize
  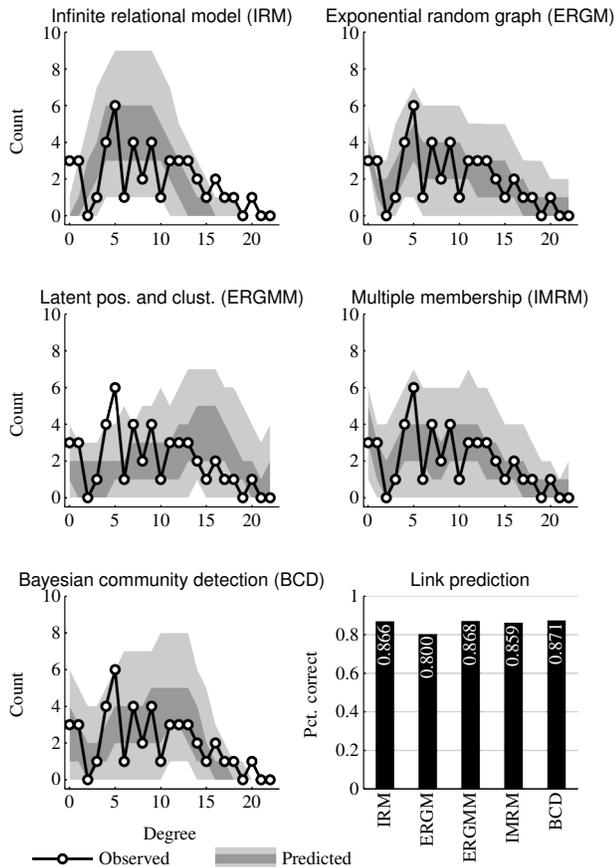
  \caption{Comparison of five network models: The plots show the
    network's observed degree distribution as well as the posterior
    predictive 95\% and 50\% intervals (shaded areas) for each of the
    models. The plot on the lower right shows the fraction of
    correctly predicted links/nonlinks when the models are trained on
    one network and used to predict links in another related network. }
  \label{fig:ComparisonWithERGM}
\end{figure}

We used the first of the two networks for training and examined the model fit by assessing the posterior predictive distribution of the node degree distribution. We fit an IRM model as well as two other non-parametric Bayesian network models, the infinite multiple membership relational model (IMRM) and the Bayesian community detection model (BCD) which are discussed further in the sequel. These models were fit using MCMC with $10,000$ rounds of Gibbs sampling where the first half of the samples were discarded for burn-in. Furthermore, we fit an exponential random graph model (ERGM) using the network statistics \emph{sociality} and \emph{gwdegree}~\cite{Morris:Handcock:Hunter:2007:JSSOBK:v24i04} as well as a latent position and cluster model (ERGMM)~\cite{Krivitsky:Handcock:2007:JSSOBK:v24i05} using a latent space of dimension four and six latent clusters (varying these parameters gave similar results). To compare how well the models fit the data we plotted the posterior predictive distribution of the degree distribution (see Figure~\ref{fig:ComparisonWithERGM}). The results show that the two most flexible models, the ERGM and the IMRM fit the data very well in terms of reproducing the degree distribution. The fit of the IRM and BCD models which are both simple latent cluster models is less good: Both models appear to underestimate the number of nodes with a high degree, i.e., employees interacting with more than 15 colleagues. The ERGMM model on the other hand appears to overestimate the number of nodes with degrees around 15--20.

Next, we compared the models' predictive performance by evaluating their ability to predict links in the second network (see Figure~\ref{fig:ComparisonWithERGM}). Here, all models except the ERGM performed on par, suggesting that the inclusion of latent variables in the model is beneficial for this task.

\section{Review of non-parametric Bayesian network models}
In the previous section we have discussed the infinite relational model, which is the most simple example of a non-parametric Bayesian latent variable model for complex network. In that model the latent variable is categorical, introducing a clustering of the network nodes; however, many other types of non-parametric Bayesian network models have been proposed in which the latent variables take other forms. Most of these can be classified as latent class, latent feature, or latent hierarchy models. In the following, we review a number of recent non-parametric Bayesian network models: We present their generative model and discuss the underlying modeling assumptions, but omit the specific details involved in inference and model checking.

\subsection{Latent class models}
In latent class models each node is assumed to belong to one class and the CRP is used a non-parametric distribution of these latent classes. The infinite relational model is the most prominent example of non-parametric latent class models for complex networks. This can be attributed to the fact that the model can capture multiple types of network structures. Contrary to other network modeling approaches such as spectral clustering~\cite{Luxburg2007} and modularity~\cite{Newman2006} groups are defined by how they interact not only internally but also externally. As such, groups are not only defined in terms of their internal properties but in particular by how they interact with the remaining parts of the network. Groups may therefore be defined as having no links between the nodes within the group as illustrated by the fourth (light blue) group of the Zachary Karate Club network in Fig.~\ref{fig:ZacharyCelegansDrugs}.

Communities in the IRM model can in turn be defined as clusters with high within-cluster density relative to their between-cluster density, interactions between groups can be accounted for by the off-diagonal elements of the $\phi$ matrix while hierarchical structures form a structured system of interaction between the elements in the $\phi$ matrix, see also Fig.~\ref{fig:OtherModels}.

The IRM model can be considered a compression of a complex network into a subgraph formed by $\phi$ that accounts for the connectivity between the components. If the number of components is the same as the number of vertices of the graph the model will recover the actual graph (when we disregard potential influences of priors) and nothing is learned in terms of structure in networks. As such the IRM model can adjust its complexity, interpolating between the full graph and the Erd\H{o}s-R\'{e}nyi graph that corresponds to an IRM model with only one component. Bayesian non-parametrics, i.e. the Chinese Restaurant Process, here admits inference over the hypothesis space encompassing all models between these two extremes in order to find plausible accounts of block structure in networks.

\subsubsection{Restrictions on cluster interactions}
Although, the IRM model is very flexible in terms of the structure it is able to account for, specialized non-parametric latent class methods have been proposed that specifically aim at extracting specific types of network structures. These models can  be characterized by the restrictions which they impose on the between-class interactions $\phi$.

In \cite{hofman2008} the $\phi$ matrix is constrained to only include two parameters, a within-group link probability $\rho_w$ and a between-group link probability $\rho_b$ such that
\begin{equation}
  \phi_{k,\ell}=\left\{\begin{array}{ll}
      \rho_w & \text{if}\ k=\ell, \\
      \rho_b & \text{otherwise}. \end{array}\right.
\end{equation}
In \cite{MMMNS_NECO2012} the within-group link probabilities are individual for each group but between-group probabilities are shared for all combinations of groups,
\begin{equation}
  \phi_{k,\ell}=\left\{\begin{array}{ll}
      \rho_\ell & \text{if}\ k=\ell,  \\
      \rho_b & \text{otherwise}. \end{array}\right.
\end{equation}

\subsubsection{Bayesian community detection}
Both of the models mentioned above are inspired by the notion of communities defined as
\begin{quotation}
``the organization of vertices in clusters, with many edges joining vertices of the same cluster and comparatively few edges joining vertices of different clusters.''~\cite{fortunato2010community}
\end{quotation}
This definition is used explicitly in \cite{MMMNS_NECO2012} forming the Bayesian Community Detection (BCD) method. The BCD is based on the following non-parametric generative model that strictly enforces community structure by constraining the diagonal elements of the $\phi$ matrix to be larger than the off-diagonal elements. The generative model for BCD is given by
\begin{align}
  \bar{z}&\sim \CRP(A),\\
  \gamma_k& \sim \Beta(\varphi,\varphi),\\
  \phi_{k,\ell}& \sim
  \left\{\begin{array}{ll}
      \Beta(a,b) & \text{if}\ k=\ell,  \\
      \BetaInc(a,b,w_{lm}) & \text{otherwise}, \end{array}\right.\\
  &\phantom{\sim}\mathrm{where}\ w_{k,\ell}=\min[\gamma_k\phi_{kk},\gamma_\ell\phi_{\ell\ell}],\\
  x_{ij}&\sim \Bernoulli(\phi_{z_iz_j}).
\end{align}
According to the model the probability of a link between communities $k$ and $\ell$ is strictly smaller than $w_{k,\ell}$ defined as the minimum over the two communities of some number $\gamma_k$ times the within-community probability $\phi_k$. This is enforced by generating the between-group probabilities according to an incomplete Beta distribution (BetaInc). The parameters $\gamma_k$ define a relative gap between link probabilities within and between communities, such that $\gamma_k=1$ says that there should be fewer links (on average) between than within, and $\gamma_k=0$ says that no links can be generated from nodes in community $k$ to other communities. The gap parameter $\gamma_{k}$ can in turn be learned from data and used to define the extend to which networks are community structured.

\subsubsection{Subset infinite relational model}
In \cite{ishiguro2012} the IRM model was extended to handle irrelevant data entries by letting these entries constitute a separate noise cluster forming the subset infinite relational model (SIRM). The generative model for SIRM can be written as
\begin{align}
r_i&\sim\Bernoulli(\lambda),\\
\phi_{k,\ell}&\sim \Beta(a,b),\\
\rho&\sim \Beta(c,d),\\
\bar{z}&\sim \CRP(A),\\
x_{ij}&\sim \Bernoulli(\phi_{z_iz_j}^{r_ir_j}\rho^{1-r_ir_j}).
\end{align}
For each node, the binary variable $r_i=0$ indicates that the node belongs to the noise cluster. For all pairs of nodes $(i,j)$ not in the noise cluster the model is identical to the IRM model; however, links between pairs of nodes of which at least one is in the noise cluster are generated with a shared probability $\rho$.

\begin{figure} 
  \centering
  \footnotesize
  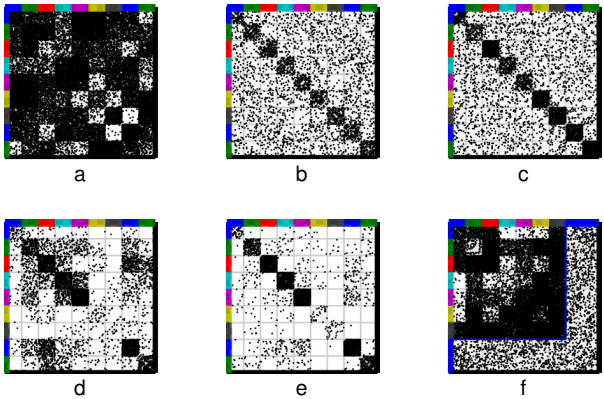
  \caption{Examples of existing latent class models. {\sffamily a)}
    The IRM model assumes arbitrary interactions between
    clusters. {\sffamily b)} The model proposed in \cite{hofman2008}
    has two parameters specifying the within group link probability
    and between group link probability.  {\sffamily c)} One of the
    models described in \cite{MMMNS_NECO2012} uses an individual
    within-cluster link probability and same between cluster link
    probability. {\sffamily d)} The BCD model of \cite{MMMNS_NECO2012}
    strictly imposes that within group link probability be larger than
    between group link probability here given for $\gamma_k=0.5$ for
    all the ten clusters. {\sffamily e)} BCD using $\gamma_k=0.1$ for
    all the ten clusters. {\sffamily f)} The SIRM model
    \cite{ishiguro2012} where nodes are divided into relevant (first 8
    clusters modelled by IRM) and irrelevant (last clusters modelled
    as noise).}
    \label{fig:LatentClassModels}
\end{figure}

All the above extensions can potentially improve on identification of structure in complex networks by substantially reducing the parameters space of the within and between group interaction matrix $\phi$ compared to the IRM model. The above extensions are illustrated in Figure~\ref{fig:LatentClassModels}.

\subsection{Latent feature models}
While latent class models restrict each node to belong to one and only one class, latent feature models endow each node with a vector of latent feature values. Exponential random graph models that embed each node in a latent feature space of fixed dimension belong to the class of latent feature models. In contrast, in non-parametric Bayesian latent feature models, the dimensionality of the latent space is learned from data to best fit the observed network. Existing non-parametric latent feature models for networks are based on the Indian Buffet Process (IBP) \cite{griffiths2006,griffiths2011}. Similarly to the CRP the IBP can be derived by starting with a finite model and considering the limit as the number of features goes to infinity. A finite set of $k=(1,\dots,K)$ binary features $z_{i,k}$ with entry 1 if node $i$ possesses feature $k$ and zero otherwise, can be generated according to
\begin{align}
  \pi_k&\sim \Beta(\alpha_k,1),\\
  z_{i,k}&\sim \Bernoulli(\pi_k).
\end{align}
Each $z_{i,k}$ is independent of all other assignments conditioned on $\pi_k$ while the $\pi_k$ are generated independently~\cite{griffiths2011}. As in the derivation of the CRP, we define $\alpha_1=\ldots=\alpha_K=A/K$ and marginalize over the nuisance parameter $\pi_k$, yielding the expression~\cite{griffiths2011}
\begin{align}
  p(\bar{Z})&=\prod_{k=1}^K\int (\prod_{i=1}^N p(z_{i,k}))p(\pi_k)d\pi_k\\
  &=\prod_{k=1}^K \frac{\alpha_k\Gamma(n_k+\alpha_k)\Gamma(N-n_k+1)}{\Gamma(N+1+\alpha_k)}.
\end{align}
Since again, as in the CRP, the labels of the features are arbitrary, we define an appropriate equivalence class for the binary matrix $Z$ by ordering the columns of the matrix from left to right according to their ``history'' $h$ in decreasing order. A history $h$ denotes one of the potential $2^{N}$ specific combinations of nodes a feature can possess enumerated according to the order of the nodes such that a feature possessed by the $n$th node contributes by a factor of $2^{N-n}$ to its history. For example in a network with 3 nodes if only node 1 and 3 possess feature $q$ the feature will have the history enumerated by $h=2^{3-1}+2^{3-3}=5$ which is greater than a feature $q^\prime$ possessed by only node 2 and 3 which has the history $h'=2^{3-2}+2^{3-3}=3$. As a result, feature $q$ will be to the left of feature $q^{\prime}$. Features which are not possessed by any nodes have $h=0$ and are ordered last. Since a permutation of the ordering of the features in $Z$ is inconsequential, we consider the equivalence class of features ordered by their history. The number of equivalent feature matrices can be computed as
\begin{align}
  \frac{K!}{\prod_{h=0}^{2^N-1}K_h!},
\end{align}
where $K_h$ is the number of features with history $h$ and $K_0$ denotes the number of features that are empty. This equivalence class is used in a similar way as when we considered the distribution over partitions in the CRP. Taking the limit yields
\begin{multline}
  \lim_{K\rightarrow \infty}p(\bar{Z})=\frac{A^{\bar
      K}\exp(-AH_N)}{\prod_{h=1}^{2^N-1}K_h!}\\\times
  \prod_{k=1}^{\bar K}\frac{\Gamma(N-n_k+1)\Gamma(n_k)}{\Gamma(N+1)},
\end{multline}
where $\bar{Z}$ denotes the left ordered equivalence class, $\bar K$ the number of non-empty features and $H_N$ denotes the $N$th harmonic number~\cite{griffiths2011}. Since this defines a distribution over an infinite size feature matrix of which only a finite subset of the features are used, the construction makes it possible to infer the number of features best suited to model the data. Compactly, we write $\bar Z \sim \IBP(A)$.

\subsubsection{Latent feature relational model}
In \cite{miller2009nonparametric} the binary matrix factorization model \cite{Meeds2007} based on an IBP is considered for network data. The following generative model embodies the latent feature relational model (LFRM)
\begin{align}
  \bar{Z}&\sim \IBP(A),\\
  \phi_{k,\ell}&\sim \Normal(0,\sigma_w^2),\\
  x_{i,j}&\sim \Bernoulli\Bigg(\sigma\bigg[\sum_{k,\ell}z_{i,k}z_{j,\ell}\phi_{k,\ell}\bigg]\Bigg),
\end{align}
where $\sigma[x]$ is a sigmoid function such as the the logit or probit. This model is inspired by the IRM in its parameterization but the model admits the nodes to belong to multiple groups, i.e., for each node to possess multiple features.

\subsubsection{Infinite latent attribute model}
In \cite{Palla2012} the infinite latent attribute model (ILAM) is proposed in which each of the nodes have a number of associated binary feature, and within each feature the nodes belong to an individual subcluster. The model can be summarized by the generative process,
\begin{align}
  \bar{Z}&\sim \IBP(A),\\
  c^{(m)} &\sim \CRP(\gamma),\\
  \phi_{k,\ell}^{(m)}&\sim \Normal(0,\sigma_w^2),\\
  x_{i,j}&\sim \Bernoulli\Bigg(\sigma\bigg[s+\sum_m
  z_{i,m}z_{j,m}\phi^{(m)}_{c_i^{(m)} c_j^{(m)}}\bigg]\Bigg).
\end{align}
For each feature $m$, the nodes that possess that feature are clustered according to a CRP. Here, $s$ is a bias term and $c^{(m)}_i$ is the cluster assignment of the $i$th node in the $m$th latent feature.

Both the LFRM and ILAM have been demonstrated to perform better than the IRM on a variety of link-prediction tasks \cite{miller2009nonparametric,Palla2012}. An important property of these models is that they allow for the membership of nodes in one group to inhibit the probability of linking to nodes in other groups as $\phi$ may include negative (i.e. antagonistic) elements. This property may indeed be an important reason for the models' superior link prediction performance over IRM.

\subsubsection{Infinite multiple-membership relational model}
In \cite{mmmns_imrm2010,MMMNS_IMRM2011} the infinite multiple-membership relational model (IMRM) was proposed. Here the probability of observing a link between vertex $i$ and $j$ is generated independently given the (multiple) groups that vertex $i$ and $j$ belongs to and their interactions $\phi$. The generative model for the IMRM is given by
\begin{align}
  \bar{Z}&\sim \IBP(A),\\
  \phi_{k,\ell}&\sim \Beta(a,b),\\ x_{i,j}&\sim\Bernoulli\Bigg(1-\prod_{k,\ell}(1-\phi_{k,\ell})^{z_{i,k}z_{j,\ell}}\Bigg).
\end{align}
If, for example, node $i$ possesses feature $k$ and node $j$ possess feature $\ell$ the quantity $\phi_{k,\ell}$ denotes the probability of a link being generated between node $i$ and $j$ on account of that pair of features. The expression $1-\prod_{k,\ell}(1-\phi_{k,\ell})^{z_{i,k}z_{j,\ell}}$ defines the probability of observing a links between vertex $i$ and $j$ as the total probability of one or more of the pairs of features possessed by the two nodes to independently generate the link. This construction is referred to as a ``noisy or process''. Notably, the IRM model is recovered when nodes belong to one and only one group.

\begin{figure} 
  \centering
  \footnotesize
  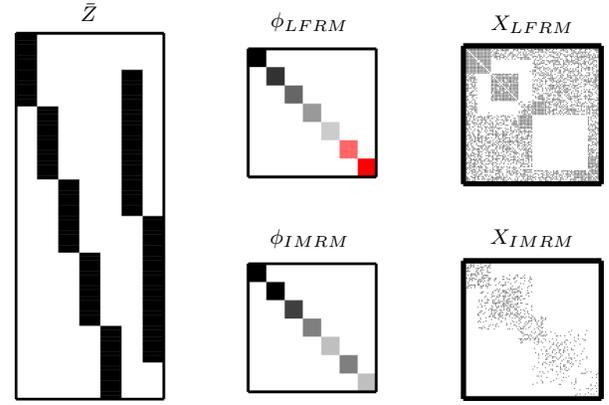
  \caption{Illustration of the LFRM and IMRM models. {\sffamily a)}
    The LFRM model assumes arbitrary interactions between latent
    features, i.e. both positive (given by the interaction within the
    first five features) and negative (given for the interaction
    within the last two features)). {\sffamily b)} The IMRM model
    assumes features act as independent causes of links such that the
    link densities monotonically increase by the number of latent
    features the nodes possess.}
    \label{fig:LatentFeatureModels}
\end{figure}

\begin{figure} 
  \centering
  \footnotesize
  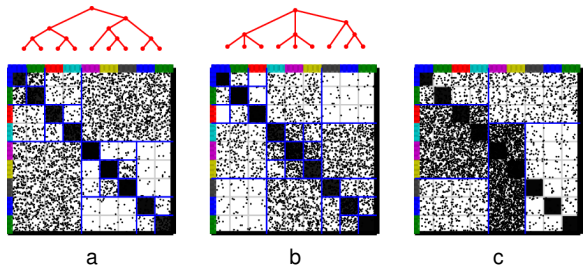
  \caption{Example of networks with hierarchical structure. {\sffamily
      a)} In the binary hierarchical relational
    models~\cite{roy2007learning,clauset2008hierarchical} link
    probability parameters are shared in a binary
    hierarchy. {\sffamily b)} In the multifurcating hierarchical
    model~\cite{HerlauEtAlCIP2012,Schmidt2012} the hierarchy at each
    level can make an arbitrary number of splits. The model is thereby
    able to infer whether or not hierarchical structure is present and
    includes the infinite relational model and the binary hierarchical
    model as special cases. {\sffamily c)} In the Mondrian
    process~\cite{RoyTeh2009a} the link probabilities are shared in a
    binary $k$-dimensional tree, corresponding to a series of axis
    aligned cuts.}
    \label{fig:OtherModels}
\end{figure}

Contrary to the LFRM and ILAM, the IMRM scales computationally in the number of observed links in the network rather than the number of potential links in the network which admits large scale analysis (see \cite{MMMNS_IMRM2011} for the details). However, scalability comes at the price of not being able to model antagonistic interactions between groups as for LFRM and ILAM. The LFRM and IMRM are illustrated in Figure~\ref{fig:LatentFeatureModels}.

\subsubsection{Latent factor models}
The IBP is useful for defining non-parametric representations of binary latent variable models and both the LFRM and ILAM can be considered non-parametric latent variable models within the exponential random graph formulation. One approach for model order selection within framework of exponential random graph models is to impose sparse priors. The IBP can here be considered a non-parametric sparse prior for latent variable modeling in general as also proposed for factor analysis in \cite{knowles2007infinite}. As such, the IBP works in a similar manner as a slab-and-spike type prior, where a feature is either present or not according to the IBP while its contribution if present can be drawn separately. This can be used to extend existing sparse latent variable models within the exponential random graph model framework to form non-parametric models.

For instance, a non-parametric version of a latent factor model~\cite{hoff_2009_cmot} can be defined by the following generative process using the IBP as a non-parametric sparsity promoting prior.
\begin{align}
  \bar{Z}&\sim \IBP(A),\\
  u_{i,k}&\sim \Normal(0,\sigma_u^2),\\
  x_{i,j}&\sim\Bernoulli\Bigg(\sigma\bigg[
  \sum_k(z_{i,k}u_{i,k})(z_{j,k}u_{j,k})\bigg]\Bigg).
\end{align}

\subsection{Latent hierarchical models}
Many complex networks are believed to be hierarchically organized such that a latent hierarchy plays an important role in accounting for the structure of the network connectivity \cite{simon1962,Ravasz2002,roy2007learning,sales2007,clauset2008hierarchical,RoyTeh2009a,Meunier2010,HerlauEtAlCIP2012}. Bayesian non-parametrics can be used to define flexible priors over all conceivable hierarchical structures and from data infer the particular hierarchical structure that is supported by the data in a similar manner as the CRP and IBP is used to infer latent clusters and features respectively.

\subsubsection{Hierarchical random graphs}
In \cite{clauset2008hierarchical} the perhaps most simple non-parametric model for hierarchical organization is proposed. This model imposes a uniform prior over all binary trees, which in the following we refer to as $\UBT$. The probability of generating a link between two nodes is defined by a parameter located at the level of their nearest common ancestor in the binary tree. A model for network with $N$ nodes thus has $N-1$ such parameters associated with each of the internal nodes in the tree. The generative model for the hierarchical random graph is given by
\begin{align}
  T&\sim \UBT(N),\\
  \phi_n&\sim \Beta(a,b),\\
  x_{i,j}&\sim \Bernoulli(\phi_{t_{i,j}}),
\end{align}
where $t_{i,j}$ denotes the index the nearest common ancestral node of vertex $i$ and $j$. In \cite{roy2007learning} a related generative model for binary hierarchies is proposed where each edge in the tree has an associated weight that defined the propensity in which the network complies with the given split.

\subsubsection{The Mondrian process}
One way to view the hierarchical random graph models is by first considering the top level of the hierarchy. Here the set of nodes is split into two partitions, and a single parameter is assigned to model the probability of observing a link between nodes in the two partitions. Next, the process continues recursively on the two partitions until each node is in a partition for itself. This framework was generalized and extended to the Mondrian process~\cite{RoyTeh2009a} which can be seen as a distribution over a $k$-dimensional tree. Used as a prior in a non-parametric Bayesian model of a bipartite network, at the top level the Mondrian process splits either of the two sets of nodes (chosen by random) into two partitions and continues this random bisectioning of the nodes until a stopping criterion is met. Parameters are then assigned to model the probability of links between each of the resulting pairs.

\subsubsection{Infinite tree-structured model}
In \cite{HerlauEtAlCIP2012} the uniform prior over binary trees of \cite{clauset2008hierarchical} where replaced by a uniform prior over multifurcating trees and the leafs of the trees rather than terminating at each vertex of the graph terminate at the levels of clusters generated from a CRP based on the following generative model
\begin{align}
  \bar{z}&\sim \CRP(A),\\
  T&\sim \UT(K_{\bar{z}}),\\
  \phi_n&\sim \Beta(a,b),\\
  x_{i,j}&\sim \Bernoulli(\phi_{t_{z_i,z_j}}).
\end{align}
Here $K_{\bar{z}}$ denotes the number of clusters in $\bar{z}$ and $\UT$ defines a uniform prior over multifurcating trees. A benefit of this model is that it can be used to detect the presence of hierarchical structure as it includes the IRM model in its hypothesis space defined by a split at the root of the tree directly into all K clusters (i.e. forming a flat hierarchy). The model of \cite{clauset2008hierarchical} can on the other hand be considered the special case where the CRP only generates singleton clusters while the tree structure is strictly binary. As the leafs terminates in clusters rather than singletons the complexity of the model is in general substantially reduced compared to the models of \cite{clauset2008hierarchical,roy2007learning,RoyTeh2009a} while the CRP defines the level at which to terminate the tree.

\subsubsection{Gibbs fragmentation trees}
In \cite{Schmidt2012} the Gibbs fragmentation tree was used as prior over multifurcating trees terminating at the vertex level of the network according to the following generative model
\begin{align}
T&\sim \GFT(\alpha,\beta),\\
\phi_n&\sim \Beta(a,b),\\
x_{i,j}&\sim \Bernoulli(\phi_{t_{z_i,z_j}}).
\end{align}
The Gibbs fragmentation tree is closely related to the two parameter nested Chinese restaurant process \cite{aldous1985} differing in explicitly accounting for the occurrence in the nested CRP of trivial non-splits. The Gibbs fragmentation tree has several attractive properties. It is i) \emph{exchangeable} in that the distribution does not depend on the labelling of the leaf nodes, ii) \emph{Markovian} in that a subtree of the full tree is in turn a Gibbs fragmentation tree, and iii) \emph{consistent} in that marginalizing over all leafs not considered in the subtree has the same distribution as only considering the Gibbs fragmentation tree of the subtree, see also \cite{mccullagh2008,Schmidt2012}. Apart from these attractive properties, the Gibbs fragmentation tree gives explicit control of the prior over multifurcating trees by its two parameters $\alpha$ and $\beta$, that makes it possible to bias the model toward deep vs. flat hierarchies. The probability of a given Gibbs fragmentation tree can be calculated using a simple recursive formula, see also~\cite{mccullagh2008,Schmidt2012}.

\subsection{Modeling side-information}
The Bayesian modeling framework readily extends to the modeling of side-information, i.e, exogenous predictors. The side-information can be used either for providing further data in support of the latent structure or directly for modeling the network links.

\paragraph{Information about latent structure}
In \cite{kemp2006learning,xu2006learning} multiple data sources were used in the IRM model to both model dyadic relationships as well as side information such that the partitioning of the nodes in the graph and the corresponding side-information available were identical.

\paragraph{Information about the network}
Instead of having the side information inform about the latent variables, it can be used to directly model the links. This approach was used in the LFRM model~\cite{miller2009nonparametric} modifying the Bernoulli likelihood function in the LFRM model according to
\begin{multline}
   x_{i,j} \sim \Bernoulli\Bigg(\sigma\bigg[\sum_{k,\ell}z_{i,k}z_{j,\ell}\phi_{k,\ell}
   +\boldsymbol{w}^\top \boldsymbol{r}_{ij}\\
   +(\boldsymbol{\gamma}^\top \boldsymbol{s}_i+a_i )
   +(\boldsymbol{\upsilon}^\top \boldsymbol{t}_j+b_j )+c
   \bigg]\Bigg),
   \end{multline}
 where $\boldsymbol{r}_{ij}$ denotes a vector of various between node similarities, $\boldsymbol{s}_i$ and $\boldsymbol{t}_i$ denotes vectors of features (i.e. side-information) for node $i$ and $j$ respectively. $\boldsymbol{w},\ \boldsymbol{\gamma},\ \boldsymbol{\upsilon}$ are parameters specifying the effect of the side-information in predicting links and $\boldsymbol{a}$, $\boldsymbol{b}$ specify node specific biases whereas $c$ is a global offset that can be used to define the overall link density. This formulation is closely related to the way in which exogenous predictors are included in the exponential random graph model.

These frameworks readily generalizes to the non-parametric latent class, feature, and hierarchical models described above and makes it possible to include all the available information when modeling complex networks. In particular including side information may improve the identification of latent structure \cite{kemp2006learning,xu2006learning} as well as the prediction of links \cite{miller2009nonparametric}.

\section{Outlook}
The non-parametric models for complex networks use latent variables to represent structure in networks. As such they can be considered extensions of the traditional exponential random graph models. The non-parametric models here provide a principled framework for inferring the number of latent classes, features, or levels of hierarchy using non-parametric distributions such as the Chinese restaurant process (CRP), Indian buffet process (IBP) and Gibbs fragmentation trees (GFT). A benefit of these non-parametric models over traditional parametric models of networks is that they can adapt to the complexity of the networks by defining an adaptive parametrization that can account for the needed level of model complexity. In addition, the Bayesian modeling approach admits a principled framework for the statistical modeling of networks and enables to take parameter uncertainty into account. In particular, the Bayesian modeling approach defines a generative process for networks which in turn can be used to simulate graphs, validate the models ability to account for network structure and predict links \cite{miller2009nonparametric,MMMNS_IMRM2011,Palla2012} while Bayesian non-parametrics bring an efficient framework for the inevitable issue of model order selection. The non-parametric Bayesian modeling of complex networks have many important challenges that are yet to be addressed. Below we outline some of these major challenges to point out some avenues of future research.

\subsection{Scalability}
Many networks are very large and efficient algorithms for inference in these large systems of millions to billions of nodes and billions to trillions of links will pose important challenges for inferring the parameters of the models. Here it is our firm belief it will be very important to focus on models that grow in complexity by the number of links rather than the sizes of the networks as well as inference procedures that can exploit distributed computing. As such, models will have to be carefully designed in order to be scalable and parallelizable. While the latent class models described all scale by the number of links the LFRM and ILAM models explicitly have to account for both links and non-links which makes them scale poorly compared to the more restricted IMRM model. Thus, flexibility here comes at the price of scalability. In particular, existing models that are scalable do not include the modeling of side-information for the direct modeling of links. Thus, future work should focus on building flexible scalable models for networks.

\subsection{Structure emerging at multiple levels}
Network structure is widely believed to emerge at multiple scales~\cite{simon1962,Ravasz2002,roy2007learning,sales2007,clauset2008hierarchical,RoyTeh2009a,Meunier2010,HerlauEtAlCIP2012}. A limitation of latent class models are that they define a given level of resolution in which structure is inferred. Whereas latent feature models can generate features defining clusters at multiple scales~\cite{Palla2012} this property can be explicitly taken into account by the latent hierarchical models. An important future challenge will be to define models that can operate at multiple scales while efficiently accounting for prominent network structure by combining ideas from the latent hierarchical models with existing latent class and feature models. This includes hierarchical models that explicitly account for community structure and models that allow for the nodes to be part of multiple groups on multiple hierarchical levels.

\subsection{Temporal evolution}
Many networks are not static but evolve over time~\cite{mucha2010,ishiguro2010,sarkar2012}. Rather than modeling snapshots of graphs as independent, taking into account the timing in which links are generated, when nodes emerge and vanish etc. potentially brings important information about the structure in these systems. To formulate non-parametric Bayesian models that can model networks exhibiting time-varying complexity, such as clusters that emerge and disappear and hierarchies that expand and contract, poses an important future challenge for the modeling of these time evolving networks.

\subsection{Generic modeling tools}
As of today non-parametric Bayesian models for complex networks often have to be implemented more or less from scratch in order to accommodate the specific structure of the networks at hand. It will be very useful in the future to develop generic modeling tools in which general non-parametric Bayesian models can be specified including how parameters are tied, various distributions invoked, and side-information incorporated. Publicly available non-parametric Bayesian software tools for complex networks that can well accommodate the needs of researchers modeling complex networks will be essential for these models to fully meet their potentials and be adopted by the many different research communities that today use models and analysis of complex network as an indispensable tool.

\subsection{Testing efficiently multiple hypotheses}
Despite the very different origin of complex networks it is widely believed generic properties exist across the domains of these systems. What are the generic properties of networks and how can they be best modelled is an important open problem that is in need of being addressed. Non-parametric Bayesian modeling forms a framework for inferring structure across multiple hypothesis. For example, the IRM model itself encompasses the hypotheses of the Erd\H{o}s-R\'{e}nyi random graph (an IRM with a single cluster) as well as the limit of the network itself (an IRM with a cluster for each node). Bayesian non-parametrics can here in general be used to infer structure across multiple hypotheses both including model order as in latent class models, feature representation as in the latent feature models, and types of hierarchies as in the latent hierarchical models.

Non-parametric Bayesian models for complex networks is emerging as a prominent modeling tool that both provides a principled framework for model order selection as well as model validation. As the non-parametric Bayesian models also can give an interpretable account of otherwise complex systems it is our firm belief these models will become essential in order to deepen our understanding of the structure and function of the many networks that surrounds us. There is no doubt the future will bring many new non-parametric Baysian models for complex networks and that these models will find important new application domains. We hope this paper will facilitate researchers to tap into the power of Bayesian non-parametric modeling of complex networks as outlined in this paper to address the major challenges we face in our effort to understand and be able to predict the behaviors of the many complex systems we are an integral part of.

\section*{Acknowledgements}
This work is funded in part by The Lundbeck Foundation.

\bibliographystyle{IEEEtranS}
\bibliography{ref}

\end{document}